\documentclass[10pt,twocolumn,letterpaper]{article}

\usepackage{iccv}
\usepackage{times}
\usepackage{epsfig}
\usepackage{graphicx}
\usepackage{amsmath}
\usepackage{amssymb}
\usepackage[utf8]{inputenc} % allow utf-8 input
\usepackage[T1]{fontenc}    % use 8-bit T1 fonts
\usepackage{booktabs}       % professional-quality tables
\usepackage{amsfonts}       % blackboard math symbols
\usepackage{nicefrac}       % compact symbols for 1/2, etc.
\usepackage{microtype}      % microtypography
\usepackage{graphicx}
\usepackage{float}
\usepackage{amsmath}
\usepackage{comment}
\usepackage{algorithm,algorithmic}
\usepackage{multicol}
\usepackage[numbers]{natbib}
\usepackage{diagbox}
\usepackage{makecell}
\usepackage{multirow}
\usepackage{adjustbox}
\usepackage{romannum}
\usepackage{color,colortbl}
\usepackage[table]{xcolor}
\usepackage[breaklinks=true,bookmarks=false]{hyperref}

\iccvfinalcopy % *** Uncomment this line for the final submission

\def\iccvPaperID{7175} % *** Enter the ICCV Paper ID here

\begin{document}

%%%%%%%%% TITLE
\title{ Beneficial Perturbation Network for Defending Adversarial Examples}

\author{Shixian Wen\\
University of Southern California\\
Los Angeles, California 90089\\
{\tt\small shixianw@usc.edu}
\and
Amanda Rios\\
University of Southern California\\
Los Angeles, California 90089\\
{\tt\small amandari@usc.edu}
\and
Laurent Itti\\
University of Southern California\\
Los Angeles, California 90089\\
{\tt\small itti@usc.edu}
}

\maketitle
%\ificcvfinal\thispagestyle{camera-ready}\fi

%%%%%%%%% ABSTRACT
\begin{abstract}
Deep neural networks can be fooled by adversarial attacks: adding carefully computed small adversarial perturbations to clean inputs can cause misclassification on state-of-the-art machine learning models. The reason is that neural networks fail to accommodate the distribution drift of the input data caused by adversarial perturbations. Here, we present a new solution - Beneficial Perturbation Network (BPN) - to defend against adversarial attacks by fixing the distribution drift. During training, BPN generates and leverages beneficial perturbations (somewhat opposite to well-known adversarial perturbations) by adding new, out-of-network biasing units. Biasing units influence the parameter space of the network, to preempt and neutralize future adversarial perturbations on input data samples. To achieve this, BPN creates reverse adversarial attacks during training, with very little cost, by recycling the training gradients already computed. Reverse attacks are captured by the biasing units, and the biases can in turn effectively defend against future adversarial examples. Reverse attacks are a shortcut, i.e., they affect the network's parameters without requiring instantiation of adversarial examples that could assist training. We provide comprehensive empirical evidence showing that 1) BPN is robust to adversarial examples and is much more running memory and computationally efficient compared to classical adversarial training. 2) BPN can defend against adversarial examples with negligible additional computation and parameter costs compared to training only on clean examples;  3) BPN hurts the accuracy on clean examples much less than classic adversarial training; 4) BPN can improve the generalization of the network 5) BPN trained only with Fast Gradient Sign Attack can generalize to defend PGD attacks. 

\end{abstract}

\begin{figure}[htb]
	\begin{center}
		\includegraphics[width=1\linewidth]{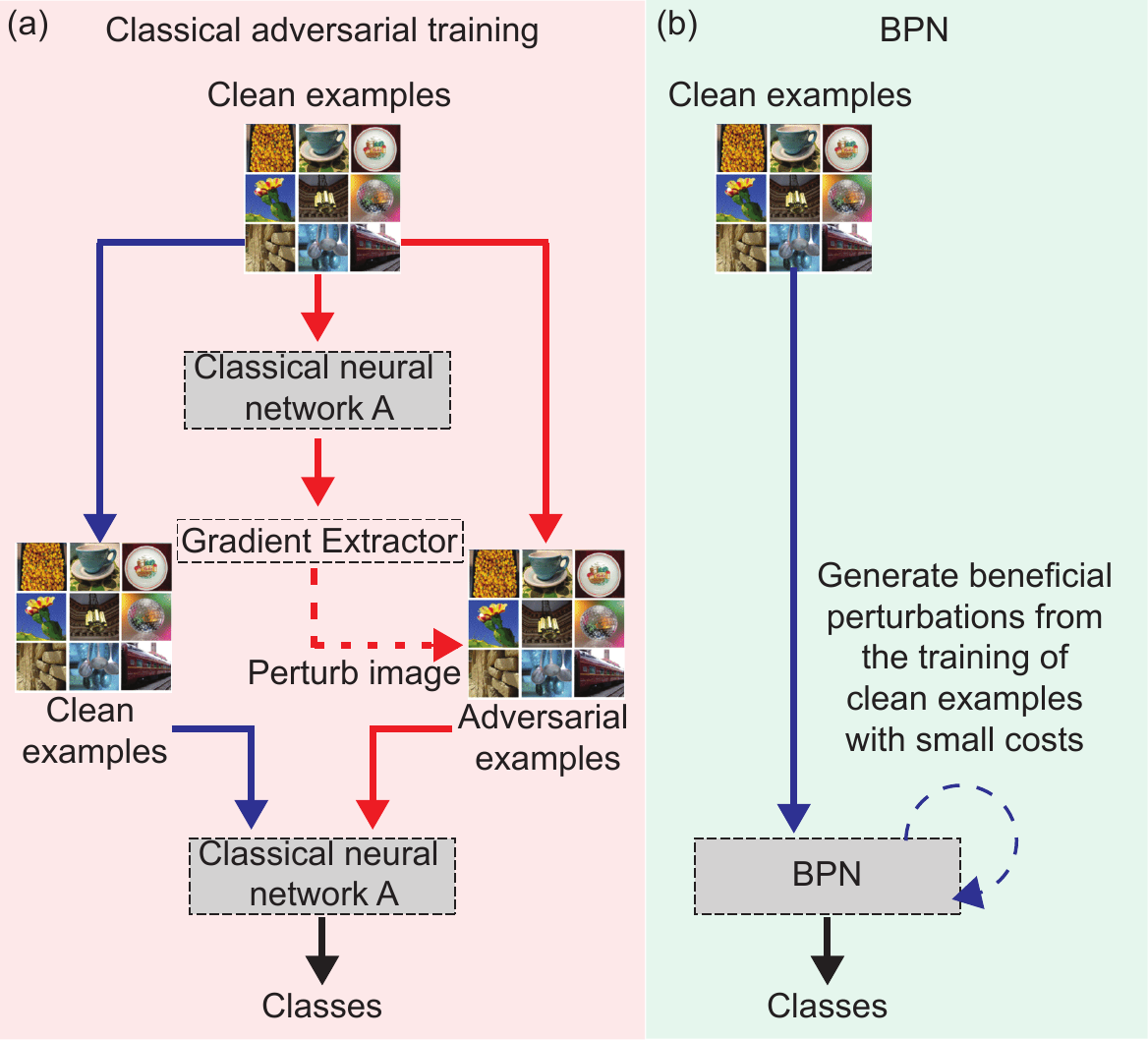}
		\caption{ Difference in training pipelines between adversarial training and BPN to defend against adversarial examples. (a) classical adversarial training has two steps: (1) Generating adversarial perturbations from corresponding clean examples and adding adversarial perturbations to the clean examples (creation of adversarial examples). (2) Training the network, usually on both clean and adversarial examples. (b) BPN creates a shortcut with only one step: training on clean examples. The feasibility of this shortcut is because BPN can generate beneficial perturbations during training on clean examples, with negligible additional costs, 0\% (0.006\%) increase for forward (backward) pass compared to training a network on clean examples. BPN is more computationally and running memory efficient compared to typical adversarial training, by recycling the gradients already computed during clean training to create beneficial biases directly in the parameter space of the network, instead of having to instantiate new perturbed images. The learned beneficial perturbations can neutralize the effects of adversarial perturbations of the data samples at test time. 
		}
		\label{fig:concepts_comparison}
	\end{center}
\end{figure}

\section{Introduction}

Neural networks have lead to a series of breakthroughs in many fields, such as image classification tasks \citep{he2016deep,chen2020improved}, and natural language processing  \citep{devlin2018bert,brown2020language}. Model performance on clean examples was the main evaluation criterion for these applications until the unveiling of weaknesses to adversarial attacks by Szegedy {\em et al.} and Biggio {\em et al.} \cite{szegedy2013intriguing,biggio2013evasion}. Neural networks were shown to be vulnerable to adversarial perturbations: carefully computed small perturbations added to legitimate clean examples  to create so-called "adversarial examples" can cause misclassification on state-of-the-art machine learning models. The reason is that adding adversarial perturbations to the input image introduces a distribution drift in the input data. Although the adversarial perturbations are often too small to be recognized by human eyes, the resulting distribution drift is sufficient to cause misclassification on machine learning models. To fix distribution drifts, a question arises: can we simulate reverse adversarial attacks during training to preempt and neutralize the effects of future adversarial perturbations?  

In this paper, we define the Beneficial Perturbations Network (BPN). BPN introduces a reverse adversarial attack to defend against adversarial examples. The key new idea is that BPN generates and leverages beneficial perturbations during training (somewhat opposite to adversarial perturbations, check Eqn.~\ref{Eqn:uptbpn2}, Eqn.~\ref{Eqn:uptbpn} and Eqn.~\ref{Eqn:uptbpn4} for detailed mathematical expressions) stored in extra, out-of-network biasing units. These units can influence the parameter space of the network, to fix distribution drifts at test time by neutralizing the effects of adversarial perturbations on data samples. The central difference between adversarial and beneficial perturbations is that, {\bf instead of adding input "noise" at test time (adversarial perturbations) calculated from other classes to force the network into misclassification, we add "noise" during training to the parameter space (beneficial perturbations), calculated from the input's own correct class to assist correct classification.}

We evaluated BPN on multiple datasets (MNIST, FashionMNIST and TinyImageNet) on three experimental scenarios:

{\bf \Romannum{1}.  Training a network on clean examples only (our main use case scenario).} This experimental scenario is preferable in the case of a modest computational budget, and where one wants to preserve clean sample accuracy while still achieving moderate robustness to adversarial examples. In this case, BPN can defend against adversarial examples with negligible additional computation costs (0\% increase for forward pass and 0.006\% for backward pass) when compared to simple clean training. As a comparison, during so-called adversarial training which is the current SOTA (see Sec.~\ref{accuracy_trade_off}), the network creates one or more adversarial examples per clean sample which means at least twice the computational power. 

{\bf \Romannum{2}. Training on adversarial examples only. }  This scenario can be used when having a modest computational budget that prioritizes robustness to adversarial examples while still wanting to preserve some amount of clean sample accuracy. When using only adversarial examples, the decision boundaries more sensitive to adversarial directions are strengthened, and this has a interesting effect of indirectly causing the model to learn some degree of clean sample representation. When compared to a classic network trained only on adversarial examples, BPN is more robust to future adversarial attacks, while also performing much better on clean samples that in fact it has never been trained on.

{\bf \Romannum{3}.  Training on both clean and adversarial examples.} This experimental scenario can be used when having abundant computational budget.
In this case, BPN is shown to be marginally superior than classical adversarial training on both clean and adversarial examples. The reason is that BPN can further improve the generalization of the network through diversification of the training set \cite{tramer2017ensemble,di2018towards,raghunathan2019adversarial,stanforth2019labels,zhang2019theoretically}. 

In addition, networks trained with classical adversarial learning have very poor generalizability to attacks that they have not been trained on. It is infeasible and expensive to introduce all unknown attack samples into the adversarial training \cite{tramer2017ensemble}.  Here,  we found experimentally that BPN trained only with FGSM  can not only defend FGSM attacks pretty well, but also generalize to defend attacks that it has never been trained on (e.g., PDG attack). 

To lay out the foundation of our approach we start by introducing the following key concepts: Sec.~\ref{accuracy_trade_off}: adversarial training; in Sec.~\ref{BPN_section}, we explain the difference between BPN and adversarial training in fixing distribution drifts of input data (Sec.~\ref{fixing_distribution_shifts}) \&  structure , updating rules (Sec.~\ref{BPN}, Sec.~\ref{BPN_loss}), computation costs (Sec.~\ref{BPN_computation}) and extension to deep convolutional network  (Sec.~\ref{BPN_extend}) for BPN.
We then present experiments (Sec.~\ref{experiments}), results (Sec.~\ref{results}) and discussion  (Sec.~\ref{discussion}).

\begin{figure*}[htb]
	\begin{center}
		\includegraphics[width=0.745\textwidth]{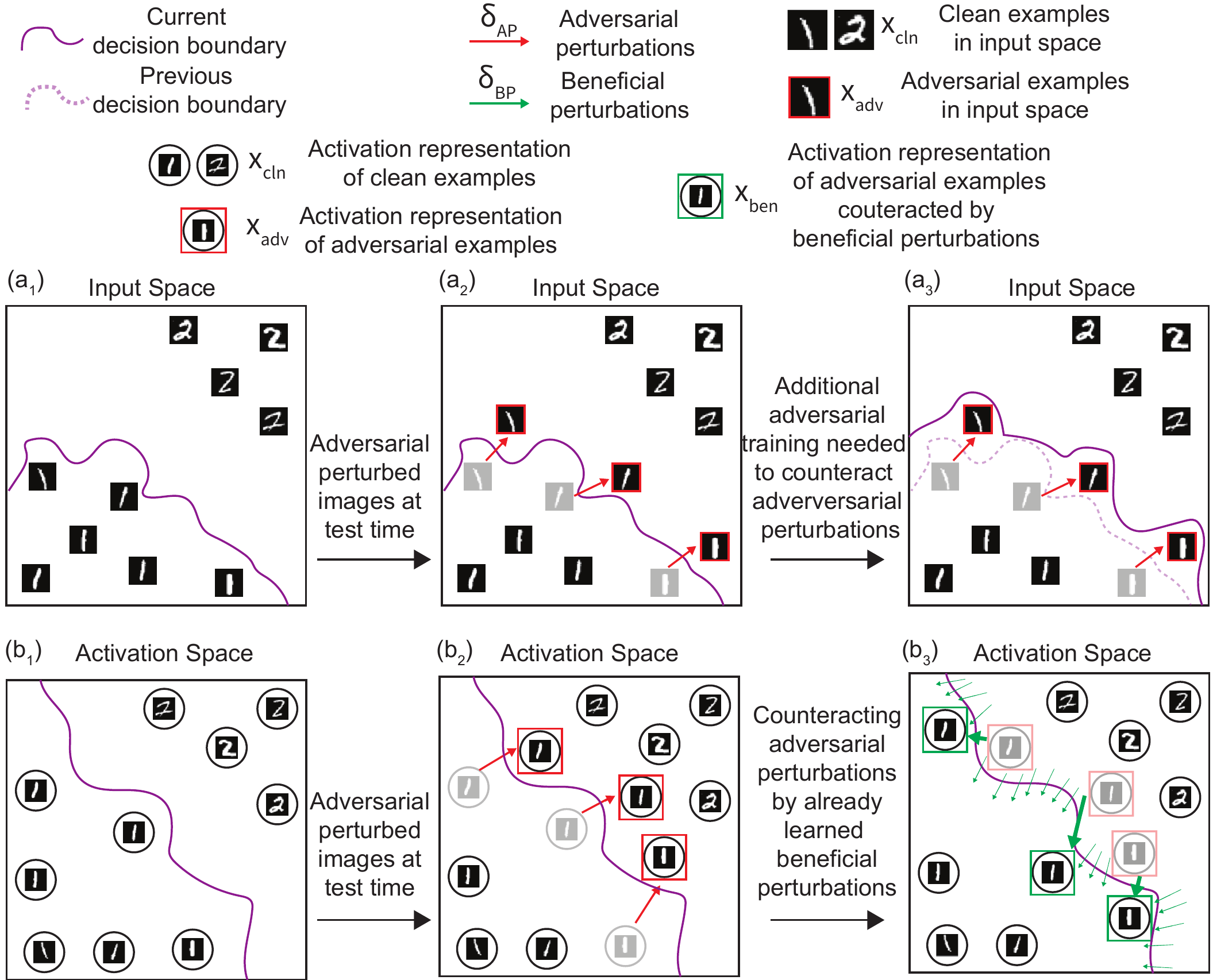}
		\caption{Difference between adversarial training ($a_1$ - $a_3$, input space) and BPN ($b_1$ - $b_3$, activation space) in fixing the data distribution drifts caused by adversarial perturbations on recognizing handwritten digits "1" versus "2". ($a_1$): After training a model on clean input images, digits "1" and "2" are separated by a purple decision boundary. ($a_2$):  Adding adversarial perturbations to test input images of digit 1 can be viewed as adding  adversarial direction vectors  (red arrows $\delta_{AP}$) to the clean (non-perturbated) input images. Such adversarial vectors cross the decision boundary, forcing the neural network into misclassification (here from digit 1 to digit 2). ($a_3$): Adversarial training: training a model on both clean and adversarial examples leads the model to learn a new decision boundary to incorporate both clean and adversarial examples, but at great computation and running memory cost. ($b_1$) and ($b_2$) are similar to ($a_1$) and ($a_2$), but they are represented in activation space. ($b_3$): BPN. Theoretically, beneficial perturbations can be seeing as the opposite of adversarial perturbations: they act to reverse the latter.  Beneficial perturbations work in activation space instead of the classical input space. In BPN, adding beneficial perturbations to the activation representation of adversarial examples corresponds to adding  beneficial direction vectors (green arrows $\delta_{BP}$) to the representations of adversarial examples of digit 1. The resulting vectors cross the decision boundary and drag the misclassified adversarial examples back to the correct classification region.
		}
		\label{fig:concepts}
	\end{center}
\end{figure*}

 \section{Related Work - Adversarial Training}
Researchers have proposed a number of adversarial defense strategies to increase the robustness of deep learning systems. Adversarial training \citep{goodfellow2014explaining,DBLP:journals/corr/HuangXSS15}, in which a network is trained on both adversarial examples  ($x_{adv}$)  and clean examples ($x_{cln}$) with class labels $y$, is perhaps the most popular defense against adversarial attacks, withstanding strong attacks. Adversarial examples are the summation of adversarial perturbations lying inside the input space ($\delta_{AP}$) and clean examples: $x_{adv}= x_{cln}+\delta_{AP}$.  Given a classifier with a classification loss function $L$ and parameters $\theta$, the objective function of adversarial training is:
\begin{equation}\smash{\displaystyle\min_{\theta}} \;  L(x_{adv},x_{cln},y;\theta) \label{Eqn:advertraining} \end{equation} 
However, adversarial training suffers from at least three difficulties:

{\bf 1. Expensive in terms of running memory and computation costs.} 
 On larger datasets, such as ImageNet, adversarial training can take multiple days on a single GPU. Kannan {\em et al.} \cite{kannan2018adversarial} used 53 P100 GPUs and Xie {\em et al.} \cite{xie2019feature} used 100 V100s for target adversarial training on ImageNet. Tramer {\em et al.} \citep{tramer2017ensemble} generate more than one adversarial examples for each clean example. These implementations require at least double the amount of running memory on GPU, to store those adversarial examples alongside the clean examples. In addition, during adversarial training, the network has to train on both clean and adversarial examples; hence, adversarial training typically requires at least twice the computation power than just training on clean examples.

{\bf 2. Accuracy trade-off.}
\label{accuracy_trade_off}
Although adversarial training can improve robustness against adversarial examples, it sometimes hurts accuracy on clean examples. Thus, there is an accuracy trade-off between the adversarial examples and clean examples \citep{di2018towards,raghunathan2019adversarial,stanforth2019labels,zhang2019theoretically}. Because most of the test data in real applications are clean examples, test accuracy on clean data should be as good as possible. Thus, this accuracy trade-off hinders the practical usefulness of adversarial training because it often ends up lowering performance on the original dataset.

{\bf 3. Impractical to foresee multiple attacks.}
Networks trained with classical adversarial learning have very poor generalizability to attacks that they have not been trained on.  
Thus, even though one might have sufficient computational resources to train a network on both adversarial and clean examples, it is infeasible and expensive to introduce all unknown attack samples into the adversarial training. For example, Tramer {\em et al.} \cite{tramer2017ensemble} proposed Ensemble Adversarial Training which can increase the diversity of adversarial perturbations in a training set by generating adversarial perturbations transferred from other models. They won the competition on Defenses against Adversarial Attacks, though again at an extraordinary computation and running memory cost.  In summary, a crucial milestone for the field of adversarial learning is achieving a model that can generalize to unseen attacks. 

\section{Beneficial Perturbation Network (BPN).}
\label{BPN_section}
Three spaces of a neural network are important: 1) The {\em input space} is the space of input data (e.g., pixels of an image); 2) the {\em parameter space} is the space of all the weights and biases of the network; 3) the {\em activation space} is the space of all outputs of all neurons in all layers in the network.

\subsection{High-level ideas - difference between BPN and adversarial training in fixing  distribution drifts of input data.}
\label{fixing_distribution_shifts}
First, how do adversarial attacks fool a neural network? For example, consider a task of recognizing handwritten digits "1" versus "2".  (Fig.~\ref{fig:concepts} $a_1$, $a_2$ in input space or their representations Fig.~\ref{fig:concepts} $b_1$, $b_2$ in activation space). Adversarial perturbations aimed at misclassifying an image of digit 1 as digit 2 may be obtained by backpropagating from the class digit 2 to the input space, following any of the available adversarial directions.  In input or activation space, adding adversarial perturbations to the input image can be viewed as adding an adversarial direction vector  (red arrows $\delta_{AP}$) to the clean (non-perturbated) input image of digit 1. The resulting vector crosses the decision boundary ({\bf input distribution drift problem}). As a consequence, adversarial perturbations can force the neural network into misclassification, here from digit 1 to digit 2 because the network failed to  accommodate the distribution drift of input data.

The primary goal of both adversarial training and BPN is to make the network more robust to adversarial attacks, but they differ in how they accomplish that.

1) In adversarial training, after training the network on both clean and adversarial examples, the network learns a new and more robust decision boundary that can accommodate input distribution drifts (Fig.~\ref{fig:concepts} $a_2$, $a_3$). In adversarial training, the decision boundary robustness is achieved via a data-driven approach. In other words, by including adversarial samples for each clean image, the decision boundary is enlarged so as to incorporate both clean and adversarial examples inside the same-label classification region. In the deployment stage, because the decision boundary is strengthened, it becomes harder to adversarially fool the network.

% In this case, by additionally including adversarial examples, these act as The robustness is achieved via a data-augmentation approach because of the inclusion of both clean and adversarial examples.

2) In BPN, to strengthen the decision boundary, we add beneficial perturbations to the activation representation of adversarial examples. In Fig.~\ref{fig:concepts} $b_2$, $b_3$, this corresponds to adding a beneficial perturbations vector (green arrows $\delta_{BP}$) to the activation representations of adversarial examples of digit 1. The resulting vector crosses the decision boundary and drags the misclassified adversarial examples back to the correct classification region ({\bf recovering from the input distribution drift caused by adversarial perturbations}). Thus, the beneficial perturbations have the effect of neutralizing the adversarial perturbations and recovering the clean examples in the activation space. In mathematical terms, as $\delta^*_{AP}$ and $\delta_{BP}$ cancel out, we have: 
\begin{equation} x^*_{cln} \approx x^*_{cln} + \delta^*_{AP} + \delta_{BP} \label{Eqn:cancelout} \end{equation}
In Eqn ~\ref{Eqn:cancelout}, $x^*_{cln}$, $\delta^*_{AP}$ are activation representations of clean examples and adversarial perturbations, respectively. As a result, BPN can achieve robusteness and correctly classify both clean and adversarial examples by training only on clean samples. Hence, unlike adversarial training, which requires several adversarial samples per clean image, BPN achieves the same goal via a much cheaper route: neutralizing future adversarial attacks with only the addition of low-cost beneficial perturbations in activation space.

\begin{figure}[h]
	\begin{center}
		\includegraphics[width=1\linewidth]{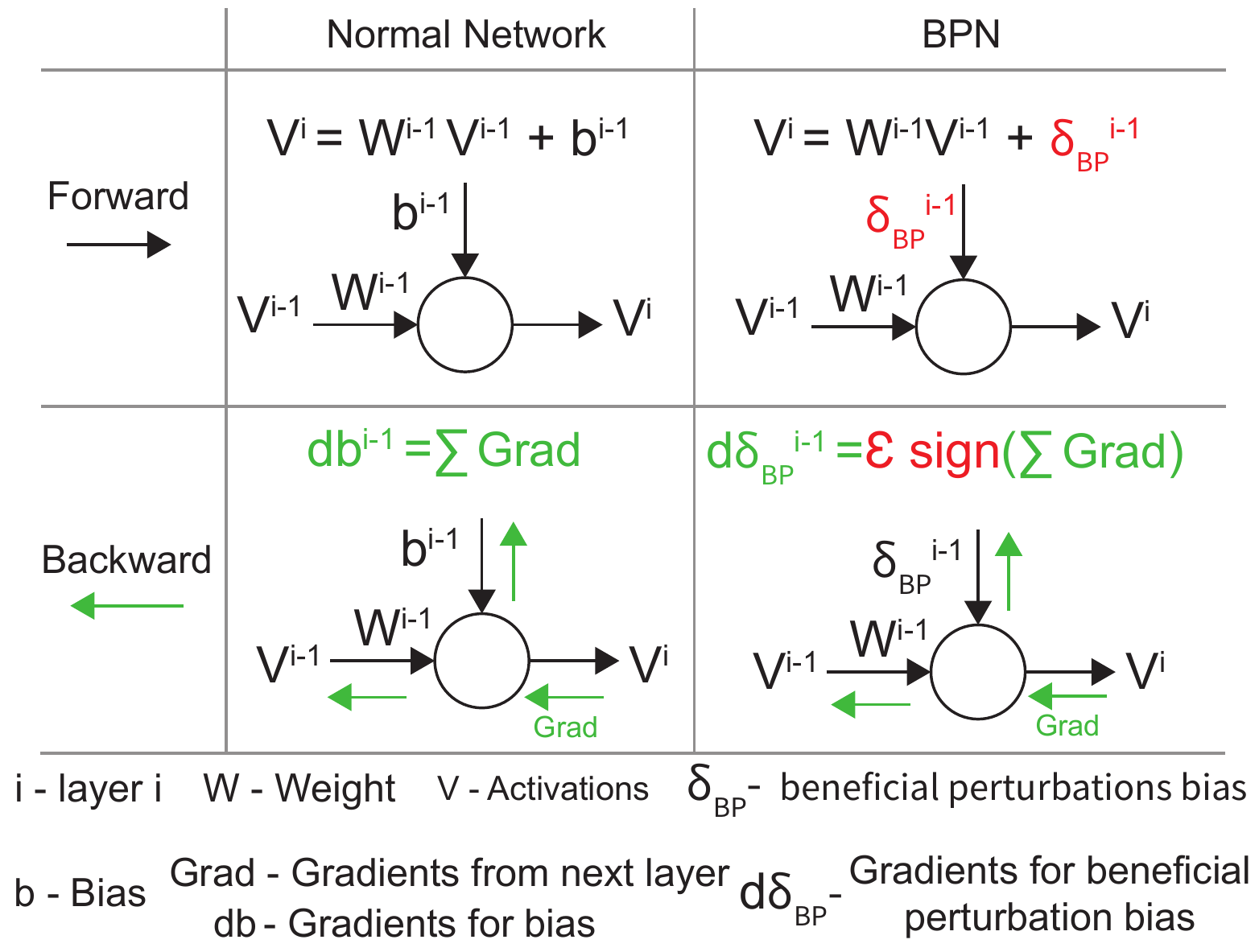}
		\caption{Structure difference between normal network (baseline) and BPN for forward (a-b) and backward pass (c-d). (a) Forward rules of normal network (baseline). (b) Forward rules of BPN. Beneficial perturbation bias ($b^{i-1}_{BP}$) is the same as normal bias ($b^{i-1}$) in forward pass. (c) Backward rules of normal network (baseline). We only demonstrated the update rules for normal bias term. (d) Backward rules of BPN. The difference is that we update the beneficial perturbations bias term using FGSM.}
		\label{fig:BPN_structure}
	\end{center}
\end{figure}

\subsection{Formulation of beneficial perturbations}
\label{BPN}
Beneficial perturbations are formulated as an additive contribution to  each layer's weighted activations (Fig.~\ref{fig:BPN_structure} b):
\begin{equation}
    V^{i+1} = W^{i}V^{i} + b_{BP}^{i}
\label{Eqn:addbpn} \end{equation}
where $W^{i}$, $V^{i}$ and $b_{BP}^{i}$ is the weight, activation and beneficial perturbation bias at layer $i$. A beneficial perturbation bias has the same structure as the normal bias term $b$    (Fig.~\ref{fig:BPN_structure} a), but it is used to store the beneficial perturbations ($\delta_{BP}$).

\subsection{Creation of reverse adversarial attack}
 Two questions arise. 1. How to create a reverse adversarial attack (beneficial perturbation). 2. How beneficial perturbation biases are trained in a way that is opposite to adversarial directions? Instead of adding input "noise" (adversarial perturbations) to the {\em input space} calculated from other classes (as in adversarial training), in BPN we  add "noise" (beneficial perturbations { $\delta_{BP}$}) to the {\em activation space} to assist in {\bf correct classification}.  These correct "noises" are learned using the gradient calculated by the input's own correct label ({ $y_{\mathbf{ true}}$}) .
 In BPN, to create beneficial perturbations, the first step is computing the adversarial direction ($d^{i}_{AP}$, Eqn.~\ref{Eqn:uptbpn5}, FGSM; \cite{goodfellow2014explaining}), using the input's own correct label ({ $y_{\mathbf{ true}}$}). However, instead of using the adversarial direction directly to create an adversarial sample, we invert the sign (${db^{i}_{BP}} = -d^{i}_{AP}$) in Eqn.~\ref{Eqn:uptbpn2} and perform  gradient descent towards to opposite direction in Eqn.~\ref{Eqn:uptbpn}, away from the adversarial vector ($d^{i}_{AP}$). This computation is repeated for each layer $i$.
 % eq 4, is the adversarial direction.
 % eq 5, we gradient descent to the opposite direction. 
\begin{equation}
        d^{i}_{AP} = \epsilon\; sign(\nabla_{b^i_{BP}} L(b^i_{BP},{\bf y_{ true}},\theta))
        \label{Eqn:uptbpn5}
\end{equation}
\begin{equation}
        db^{i}_{BP} = -d^{i}_{AP}
        \label{Eqn:uptbpn2}
\end{equation}
% \begin{equation}
%         db^{i}_{BP} = \epsilon\; sign(\nabla_{b^i_{BP}} L(b^i_{BP},{\bf y_{ true}},\theta))
%         \label{Eqn:uptbpn2}
% \end{equation}
\begin{equation}
    b^i_{BP} = b^i_{BP} - \eta \; (db^{i}_{BP})
\label{Eqn:uptbpn} \end{equation}
where $\eta$ is the learning rate, $d^{i}_{AP}$ is the adversarial perturbation, $b^i_{BP}$ is the beneficial perturbation bias, $db^{i}_{BP}$ is the gradient for beneficial perturbation bias, $\epsilon$ is a hyperparameter that decides how far we go towards the Fast Gradient Sign direction, $y_{true}$ is the {\bf true} label (input's own correct class) and $\theta$ are the parameters of the neural network.

The creation of beneficial perturbations only requires the true labels and the same gradients as we would normally have to train a Vanilla deep neural network. Thus, we can generate the gradient for beneficial perturbations at layer $i$ by recycling the computed gradients while we are training the network on classic gradient descent (Fig.~\ref{fig:BPN_structure}). Therefore we can consolidate Eqn.~\ref{Eqn:uptbpn2} and Eqn.~\ref{Eqn:uptbpn} into Eqn.~\ref{Eqn:uptbpn4}:
% \begin{equation}
%     db^{i}_{BP} = \epsilon \; sign(\sum Grad)
%     \label{Eqn:uptbpn3}
% \end{equation}
\begin{equation}
    b^i_{BP} = b^i_{BP} - \eta \; (-\epsilon \; sign(\sum Grad_{SGD}))
\label{Eqn:uptbpn4} \end{equation}
where, $db^{i}_{BP}$ is the gradient for beneficial perturbations bias and $Grad_{SGD}$ is the gradient calculated by classical stochastic gradient descent from next layer $i+1$, $\epsilon$ is same as Eqn.~\ref{Eqn:uptbpn2}. 
\subsection{Computation costs}
\label{BPN_computation}
To generate beneficial perturbations, we do not introduce any extra computation costs beyond FGSM. The forward (backward) pass computation costs of BPN are only 0.00\% (0.006\%) FLOPS more than the costs of the base network trained on clean examples only (Tab.~\ref{tab:BPN_computation_costs}).  BPN creates a shortcut (Fig.~\ref{fig:concepts_comparison}) in the training process that simulates a reverse adversarial attack in activation space. Because of this shortcut, we don't have to instantiate the adversarial examples in the original image space (input space) like in standard adversarial training. This advantage saves a lot of time and enables BPN to defend against adversarial examples robustly, even after only using clean examples for training.
\begin{table}[htb]
    \centering
        \caption{Computation costs of BPN trained on clean examples compared to a classical network trained on clean examples on RestNet-50. For forward (backward) pass, The computation costs of BPN are 0.00\% (0.006\%) FLOPS more than the classical network.} 
    \begin{adjustbox}{width = 1\linewidth}
    \begin{tabular}{c|c|c}
         \hline
         \diagbox{network}{\makecell{Computation\\ cost}}& \makecell{Forward \\(FLOPS) } & \makecell{Backward\\(FLOPS)} \\
         \hline
         \makecell{Classical Network\\ (ResNet-50)} & 51,112,224& 51,112,225\\
         \hline
        BPN (ResNet-50) & 51,112,224& 51,115,321\\
        \hline
    \end{tabular}
    \end{adjustbox}

    \label{tab:BPN_computation_costs}
\end{table}

\subsection{Loss function, forward and backward rules}
\label{BPN_loss}
We present the loss function, forward and backward rules for BPN -

\underline {Forward rules:} \;\; $ \mathbf{V}^{i+1} = \mathbf{W}^{i} \mathbf{V}^{i} + \mathbf{b}_{BP}^{i}$ \;\;\;

\;\;\; where $\mathbf{W}^{i}$, $\mathbf{V}^{i}$ and $\mathbf{b}_{BP}^{i}$ are the weight, activation and beneficial perturbation bias at layer $i$ respectively.

\underline {Backward rules:}

 Minimize loss function: {\small $\begin{aligned} \mathop{\min}_{ \,\mathbf{b}^i_{BP},\mathbf{W}^i}\end{aligned}$
$ L(f(\mathbf{x},\mathbf{y}_{ true}); \mathbf{b}^i_{BP},\mathbf{W}^i)$}

\;\;\; $d\mathbf{W}^i = \mathbf{Grad_{SGD}}\cdot((\mathbf{V}^{i})^T)$

\;\;\; $\mathbf{dV^{i}} = (\mathbf{W}^{i})^T \cdot (\mathbf{Grad_{SGD}})$

\;\;\; $\mathbf{db}^i_{BP} =  -\epsilon\; sign(\sum_{j} {Grad_{SGD}}_j)$ 

{$\; \; \; \; \; \; \; \; \;$ \color{green} // adversarial direction caculated by FGSM attack }

\;\;\; where $\mathbf{d*}^{i}$is the gradient for *, $\mathbf{Grad_{SGD}}$ is the gradient calculated by stochastic gradient descent from the next layer, $\mathbf{x}$ and $\mathbf{y}_{true}$ are the image inputs and its true label. $f$ and $L$ is the network model and cross entropy loss, $j$ is an iterator over the first dimension of $\mathbf{Grad_{SGD}}$, other notations are same as notations in forward rules.
\begin{figure*}[htb]
	\begin{center}
		\includegraphics[width=.75\linewidth]{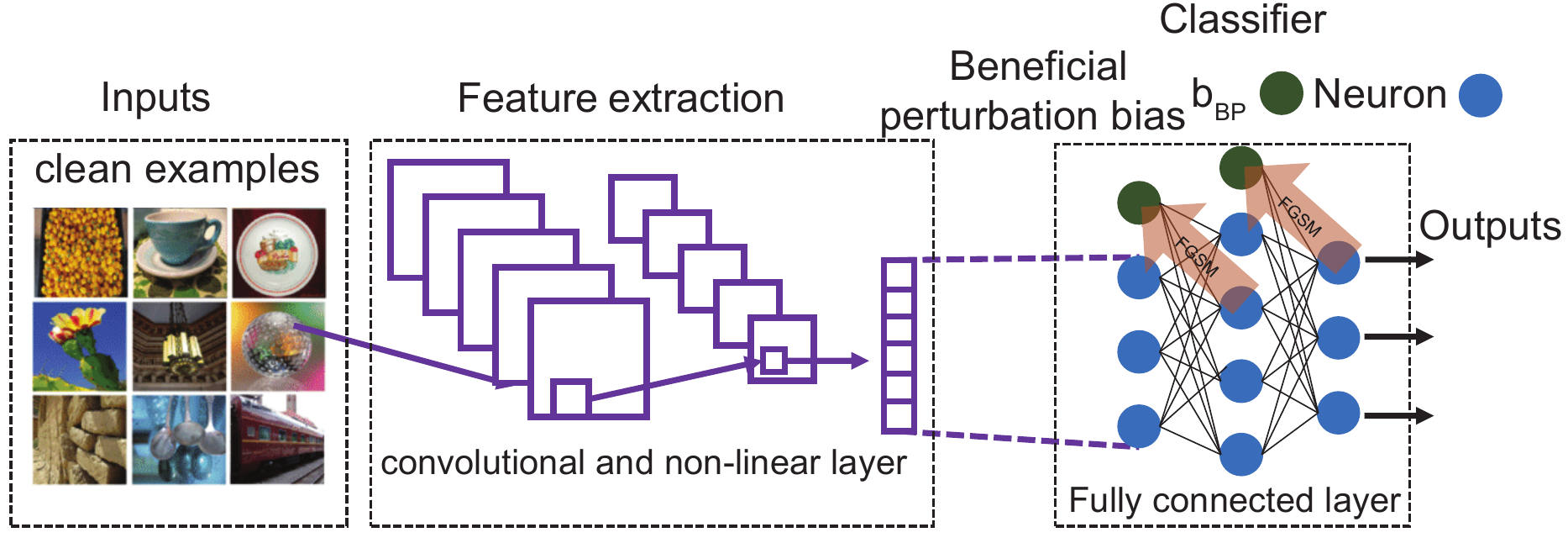}
		\caption{BPN extension to deep convolutional neural network. Deep convolutional neural network are made with two parts: feature extraction part (convolutional and non-linear layers) and classifier part (fully connected layer). We introduce beneficial perturbation bias (replace the normal bias term) to the last few fully connected layers of the deep convolutional network and update them using FGSM.}
		\label{fig:CNN_network_structure}
	\end{center}
\end{figure*}

\subsection{Extending BPN to deep convolutional networks}
\label{BPN_extend}
Most deep convolutional neural networks are made with two parts: a feature extraction part (convolutional and non-linear layers) and a classifier (fully connected layers). Here, we introduce beneficial perturbations bias ($b_{BP}$) to the last few fully connected layers of the deep convolutional network, replacing the normal bias term (Fig.~\ref{fig:CNN_network_structure}). We use FGSM (Eqn.~\ref{Eqn:uptbpn4}) to update those beneficial perturbation biases.  

%In summary, through the training of clean examples, BPN generates beneficial perturbations as biases within the last few fully connected layer of the deep neural network. BPN leverages these beneficial perturbations to defend against future adversarial examples by neutralizing the effects of adversarial perturbations in the datasets. In addition, if  BPN is trained on adversarial examples alone or on a combination of adversarial and clean examples, the neutralization can diversify the training set by converting adversarial examples to clean examples. As a consequence, the diversification further improves the generalization of the BPN. The generalization eases the difficulties of accuracy trade-off and impracticability to foresee multiple attacks.

\section{Experiments}

\label{experiments}

\subsection{Datasets}

{\bf MNIST.} MNIST \citep{lecun1998gradient} is a dataset with handwritten digits, with a training set of 60,000 examples, and a test set of 10,000 examples. 

{\bf FashionMNIST.} FashionMNIST \citep{xiao2017fashion} is a dataset of article images, with a training set of 60,000 examples, and a test set of 10,000 examples. 

{\bf TinyImageNet.}  TinyImageNet  is a subset of the ImageNet \citep{imagenet_cvpr09} - a large visual dataset. TinyImageNet consists of 200 classes and has a training set of 100k examples, and a test set of 10k examples. %Following \citep{DBLP:journals/corr/HeZRS15,DBLP:journals/corr/HeZR016,DBLP:journals/corr/HuangLW16a}, we report classification errors on the validation set.

\subsection{Network structure}
{\bf For MNIST and FashionMNIST (LeNet). } We use the convolutional and non-linear layers of LeNet as feature extraction part \citep{lecun1998gradient} (classical LeNet). Then, for the classifier part, we create our version of LeNet (LeNet with beneficial perturbation bias) by adding beneficial perturbation biases into the fully connected layers, replacing the normal biases.

{\bf For TinyImageNet (ResNet-18).} We use the convolutional and non-linear layers of ResNet-18 \citep{DBLP:journals/corr/HeZRS15} for the feature extraction part (classical ResNet-18). Then, we use three fully connected layers with 1028 hidden units as a classifier.  We create our version of ResNet-18 (ResNet-18 with beneficial perturbation bias) by adding beneficial perturbation biases into the fully connected layers, replacing the normal biases. We trained the BPN (ResNet-18) with 5000 epochs on TinyImageNet. 

% \subsection{Generation of test adversarial examples}
%  After training the BPN, adversarial examples are generated from the trained BPN against itself. For clean model, adversarial examples are generated from the trained clean model against itself.

\subsection{Various attack methods}
To demonstrate how BPN can successfully defend against adversarial attacks, we used the advertorch toolbox \citep{ding2019advertorch} to generate adversarial examples and tested our BPN structure on adversarial examples generated from various attack methods. We employ only white box attacks, where the attacker has access to the model's parameters. All adversarial examples were generated directly from the same model that they attacked (e.g., BPN generates adversarial examples against itself and likewise the classical network generates adversarial examples against itself). We tested three adversarial attacks:

{\bf{(1)}} PGD Linf \citep{madry2017towards}: Projected Gradient Descend Attack with order = Linf. 

{\bf{(2)}} PGD  L2 \citep{madry2017towards}: Projected Gradient Descend Attack with order = L2.

{\bf{(3)}} FGSM \citep{goodfellow2014explaining}: One step fast gradient sign method.

\section{Results}
\label{results}
\subsection{In scenario \Romannum{1}: BPN can defend adversarial examples with additional negligible computational costs}
When the neural network can only be trained on clean examples because of modest computation budgets, the biggest achievement of BPN is that it can defend against adversarial examples with only a low computational overhead. BPN achieves much better test accuracy on adversarial examples than a classical  network (baseline, Tab.~\ref{Tab:clnexm}, MNIST: 98.88\% vs. 18.08\%, FashionMNIST: 54.07\% vs. 11.87\%, TinyImageNet: 53.29\% vs. 1.45\% ). Thus, for companies with modest computation resources, BPN can help a system achieve moderate robustness against adversarial examples, while only introducing additional negligible computation costs. For example, on FashionMNIST, our method only uses 59\% training time compared to adversarial training that uses just one adversarial example per clean example, saving 43.51 minutes training time for 500 training epochs on an NVIDIA Tesla-V100 platform. Just for reference, the classical Stochastic gradient descent clean training would be at 50\% compared to the same adversarial training framework. Mathematically, our method only introduces 0.006\% cost compared to clean training because we only introduced one Sign and one multiplication operation for each fully connected layer Tab.~\ref{tab:BPN_computation_costs} (which means BPN approach a 50\% training time compared to adversarial training). However, in our current implementation, we incur an extra 9\% cost because we created a custom layer in Pytorch framework that introduces a lot of overhead. This can be greatly improved if the custom layers are incorporated into the Pytorch framework with a C++ implementation. The computational savings would be huge on a larger dataset such as Imagenet \citep{deng2009imagenet}).

\definecolor{LightCyan}{rgb}{0.88,1,1}

\begin{table}[h]
\centering
\caption{Scenario \Romannum{1} Training on only clean examples for both BPN and classical network (CN) because of modest computational budget. Testing on clean examples (Cln Ex) and adversarial examples (Adv Ex) (generated by FGSM, $\epsilon$ = 0.3 for MNIST, FashionMNIST and TinyImageNet). CN does poorly on adversarial examples. While, BPN can successfully defend adversarial examples (emphasized by light cyan background).}
\begin{adjustbox}{width = 1\linewidth}
\begin{tabular}{*5c}
\toprule
\multicolumn{2}{c}{\makecell {Datasets \& \\  network  structure}} &  \makecell {MNIST \\  LeNet} & \makecell {FasMNIST \\  LeNet}   & \makecell {TinyImageNet \\  ResNet-18}\\
\hline
\multirow{2}{*}{Cln Ex} & BPN & \bf{99.17} & \bf{89.53} & {57.55} \\ & CN & 99.01 & 89.17 & \bf{64.30}\\
\hline
%\cellcolor{LightCyan}
%\rowcolor{LightCyan}
 \multirow{2}{*}   {Adv Ex} & \cellcolor{LightCyan} BPN & \cellcolor{LightCyan}\bf{98.88} & \cellcolor{LightCyan}\bf{54.07} &\cellcolor{LightCyan} \bf{53.29} 
\\ & \cellcolor{LightCyan} CN & \cellcolor{LightCyan}18.08 & \cellcolor{LightCyan}11.87 &\cellcolor{LightCyan} 1.45\\
\bottomrule
\end{tabular}
\end{adjustbox}
\label{Tab:clnexm}
\end{table}

\subsection{In scenario \Romannum{2}: BPN can alleviate the decay of clean sample accuracy}
When the neural network can only be trained on adversarial examples because of modest computation power, BPN becomes naturally more robust to adversarial examples but it also performs much better on clean samples that in fact it has never been trained on. With a classical neural network (adversarial-only network), training it on only adversarial examples produces a high test accuracy on adversarial examples but it hurts test accuracy on clean examples (which it never saw) (Tab.~\ref{Tab:advexm}). The clean sample accuracy this adversarial-only network decreases from 99.01\%, 89.17\%, 64.30\% (upper-bound classical network trained only clean images to 95.54\%, 65,64\%, 18.67\%  (Tab.~\ref{Tab:advexm}) for MNIST, FashionMNIST and ImageNet datasets resectively. Compared to this adversarial-only network, BPN not only achieves better test accuracy on adversarial examples (Tab.~\ref{Tab:advexm}, MNIST: 99.27\%, FashionMNIST: 92.07\%, TinyImageNet: 79.92\%), but also achieves a better test accuracy on clean examples (Tab.~\ref{Tab:advexm}, MNIST: 97.32\%, FashionMNIST: 71.54\%, TinyImageNet: 20.69\%). This accuracy on clean examples is still worse than the upper bound (classic network trained only on clean examples), but it is much better than the accuracy of the classic network trained only with adversarial images. The reason is that beneficial perturbations would convert some adversarial examples into clean examples because of the neutralization (Eqn.~\ref{Eqn:cancelout}) effect. As a consequence, the increased diversity of the clean examples improves the generalization of BPN.

% When using only adversarial examples, the decision boundaries more sensitive to adversarial directions are strengthened and this has a interesting effectof indirectly causing the model to learn some degree of cleansample representation.

% For instance, when compared to an upper bound of training on only clean samples, 

% the test accuracy on clean examples the adversarial-only network B 

% In contrast, a classical network B trained only on clean examples, compared network A, the test accuracy on clean examples  decreases from 99.01\%, 89.17\%, 64.30\% (Tab.~\ref{Tab:clnexm}) to 95.54\%, 65,64\%, 18.67\% (Tab.~\ref{Tab:advexm}) for MNIST, FashionMNIST and ImageNet datasets resectively. 

%In comparison to classical network B, by training  BPN only on adversarial examples,
%#########################################################################
%I am not sure whether should I cite the other paper to mention this or not.
% since I am don't want to spend too much space here to explain the "error tolerance"
% I think this is just a consequence of increasing the "error tolerance" of the network in the boundaries by the effect of BPN. Similar to what you argued in your paper for continual learning. You increase the confidence region. Maybe you can say that again here inside this new context.
% Also, I think it may also be because the gradient training for the network is still classic SGD and thus the influence of the BDs remains more compartimentalized and less harmful to clean examples. 
%########################################################################

\begin{table}[h]
\centering
\caption{Scenario \Romannum{2}: Training on only adversarial examples for both BPN and classical network (CN) with modest computational budget. Testing on clean examples (Cln Ex) and adversarial examples (Adv Ex) (generated by FGSM, $\epsilon$ = 0.3 for MNIST , FashionMNIST and TinyImageNet). BPN can achieve a better classification accuracy on clean examples than  CN (emphasized by light cyan background).}
\begin{adjustbox}{width = 1\linewidth}
\begin{tabular}{*5c}
\toprule
\multicolumn{2}{c}{\makecell {Datasets \& \\  network  structure}} &  \makecell {MNIST \\  LeNet} & \makecell {FasMNIST \\  LeNet}   & \makecell {TinyImageNet \\  ResNet-18}\\
\hline
\multirow{2}{*}{Cln Ex} &\cellcolor{LightCyan} BPN & \cellcolor{LightCyan}\bf{97.32} &\cellcolor{LightCyan} \bf{71.54} &\cellcolor{LightCyan} \bf{20.69} \\ & \cellcolor{LightCyan}CN & \cellcolor{LightCyan}95.54 &\cellcolor{LightCyan} 65.64 &\cellcolor{LightCyan}18.67\\
\hline
\multirow{2}{*}{Adv Ex} & BPN & \bf{99.27} & \bf{92.07} & \bf{79.92} \\ & CN & 99.01 & 91.49 & 68.52\\
\bottomrule
\end{tabular}
\end{adjustbox}
\label{Tab:advexm}
\end{table}

\subsection{In scenario \Romannum{3}: BPN can improve  generalization through  diversification of the training set }
 When the neural network can  be trained on both clean and adversarial examples because of abundant computation power, BPN is shown to be marginally superior than classical adversarial training (Tab.~\ref{Tab:advclnexm}). The reason is that BPN can further improve the generalization of the network by diversifying the training set with the neutralization (Eqn.~\ref{Eqn:cancelout}) effect. BPN can achieve slightly higher accuracy on clean examples than the classical network  (Tab.~\ref{Tab:advclnexm} MNIST 99.13\% vs. 99.09\%, FashionMNIST 89.65\% vs. 89.49\%, TinyImageNet 66.84\% vs. 66.56\% ). In addition, BPN can achieve higher accuracy on adversarial examples than the classical Network  MNIST 97.62\% vs. 97.01\%, FashionMNIST 95.39\% vs. 94.98\%, TinyImageNet 88.16\% vs. 85.75\%). However, we should normally avoid this scenario because training on both clean and adversarial examples is expensive in terms of running memory and computation costs. In addition and most importantly, it is infeasible and expensive to introduce all unknown attack samples into the adversarial training \citep{tramer2017ensemble} (this would require generating at least one attack-specific adversarial image per attack). Instead of this data-driven data-augmentation approach, our model should be generalizable to unknown attacks, see our results Sec.~\ref{generalizable}.
 
\begin{table}[h]
\centering
\caption{Scenario \Romannum{3}: Training on both clean and adversarial examples for both BPN and classical network (CN) with abundant computational power. Testing on clean examples (Cln Ex) and adversarial examples (Adv Ex) (generated by FGSM, $\epsilon$ = 0.3 for MNIST, FashionMNIST and ImageNet). BPN is marginally superior to CN. However, we should avoid this scenario because it is expensive in terms of running memory computation costs and infeasible to introduce to all unknown attack samples into the adversarial training.}
\begin{adjustbox}{width = 1\linewidth}
\begin{tabular}{*5c}
\toprule
\multicolumn{2}{c}{\makecell {Datasets \& \\  network  structure}} &  \makecell {MNIST \\  LeNet} & \makecell {FasMNIST \\  LeNet}   & \makecell {TinyImageNet \\  ResNet-18}\\
\hline
\multirow{2}{*}{Cln Ex} & BPN & \bf{99.13} & \bf{89.65} & \bf{66.84} \\ & CN & 99.09 & 89.49 & 66.56\\
\hline
\multirow{2}{*}{Adv Ex} & BPN & \bf{97.62} & \bf{95.39} & \bf{88.16} \\ & CN & 97.01 & 94.98 & 85.75\\
\bottomrule
\end{tabular}
\end{adjustbox}
\label{Tab:advclnexm}
\end{table}

% \subsection{Influence of adversarial perturbation budget}
%  The higher the adversarial perturbation budget, the higher the chance it can successfully attack a neural network. However, attacks with higher adversarial perturbation budgets are easier to detect by a program or by humans. For example, $\epsilon =0.3$ (Fig.~\ref{fig:adversarial_bugets}a) represents very high noise, which makes FashionMNIST images difficult to classify, even by humans. But the distribution differences between the adversarial examples and clean examples are so large that they can be easily captured by defense programs. Thus, $\epsilon \leq 0.15$  is a good attack since the differences caused by adversarial perturbations are too small to be detected by most defense programs. For small adversarial perturbations (Fig.~\ref{fig:adversarial_bugets}b $\epsilon \leq 0.15$), by just training on clean images, BPN achieves moderate robustness against adversarial examples with negligible costs. Thus, it is really beneficial to adapt our method for companies with modest computation power, who still want to achieve moderate robustness against adversarial examples.
% \begin{figure}[htb]
% 	\begin{center}
% 		\includegraphics[width=0.7\linewidth]{Influence_of_adversarial_perturbation_budgets.pdf}
% 		\caption{(a) Adversarial example with high adversarial perturbation budget ($\epsilon$ = 0.3). (b) Test accuracy on adversarial examples after training BPN only on clean examples from MNIST (Blue) or FashionMNIST (Orange) datasets. }
% 		\label{fig:adversarial_bugets}
% 	\end{center}
% \end{figure}

\begin{table}[htb]
    \centering
    \caption{BPN can generalize to unseen attacks. We go back to  scenario \Romannum{1} where both BPN and classical network (CN) are trained on clean examples of MNIST and TinyImageNet because of modest computation budget. Testing on adversarial examples generated by FGSM attack ($\epsilon$ = 0.3) and PGD attack ($\epsilon$ = 0.3, number of iteration = 40, attack step size = 0.01, random initialization = True, order = Linf or L2). BPN trained with FGSM is not only robust to FGSM attack, but also can successfully generalize to defend PGD attacks that it has never been trained on.}
    \begin{adjustbox}{width = \linewidth}
    \begin{tabular}{|c|c|c|c|c|c|c|c|}
    \hline
    \multicolumn{2}{|c|}{\diagbox{\makecell{Dataset \\ \& network}}{Attacks}} & \makecell{FGSM} & \makecell{PGD Linf}  &  {PGD L2}  \\
    \hline
    \multirow{2}{*}{MNIST} & BPN & \bf{98.35} & \bf{95.41}   & \bf{98.52} \\ & CN & 17.53 & 2.18 & 97.26   \\
    \hline
    
    \multirow{2}{*}{TinyImageNet} & BPN & \bf{52.39} & \bf{44.37} &\bf{16.23} 
    \\ & CN & 1.29 & 0.00 & 15.11   \\
    \hline
    
    \end{tabular}
    \end{adjustbox}
    \label{Tab:attacks}
\end{table}

\subsection{BPN can generalize to unseen attacks that it has never been trained on}

\label{generalizable}
Here, we go back to scenario \Romannum{1} to test the generalization ability of BPN.  When the neural network can only be trained on clean examples because of modest computational budget, we trained BPN on clean examples with FGSM and tested on adversarial examples generated by various attack methods (e.g., FGSM and PGD attacks).

In standard adversarial training, a model can only defend against the kind of adversarial examples that it has been trained on \citep{goodfellow6572explaining,kannan2018adversarial} (e.g., a model trained on adversarial examples created by FGSM attack can only defend FGSM attack and would fail to defend other attacks such as PGD attack). Similarly, if BPN is updated only by the adversarial direction generated by FGSM attack, the expectation would be that the BPN can only defend FGSM attack. However, from Tab.~\ref{Tab:attacks}, we found that BPN trained only with FGSM  can not only  defend FGSM attacks pretty well, but also can generalize to improve the robustness against even harder attacks that it has never been trained on (e.g., PGD attack in Tab.~5). This feature alone has an edge compared to standard adversarial training since it is infeasible to introduce all unknown attack samples into adversarial training.

The influence of the adversarial perturbation budget is discussed in the Supplementary section. We discuss more possibilities to further improve the robustness and generalization of BPN by exploring more structures and training procedures of BPN in the Supplementary section.

\section{Discussion}

\label{discussion}

We proposed a new solution for defending against adversarial examples, which we refer to as Beneficial Perturbation Network (BPN). BPN, for the first time, leverages the beneficial perturbations (opposite to well-known adversarial perturbations) to counteract the effects of adversarial perturbations input data. Compared to adversarial training, this approach introduces four  main advantages - We demonstrated that {\bf (1)}  BPN can effectively defend adversarial examples with negligible additional running memory and computation costs; {\bf (2)} BPN can alleviate the accuracy trade-off - hurts the accuracy on clean examples less than classical adversarial training; {\bf(3)} The increased diversity of the training set can improve generalization of the network; {\bf (4)} Compared to adversarial training that can only defend the kind of adversarial examples that is has been trained on. We found experimentally that BPN has the ability to generalize to unseen attacks that it has never been trained on.
\subsection{Intriguing property of beneficial perturbations}

We suggest that the intriguing property of the beneficial perturbations that neutralize the effects of adversarial examples might come from the property of adversarial subspaces. Following the adversarial direction, such as by using the fast gradient sign method (FGSD) \cite{goodfellow6572explaining}, can help in generating adversarial examples that span a continuous subspace of large dimensionality (adversarial subspace). Because of “excessive linearity” in many neural networks \cite{tramer2017space} \cite{goodfellow2016}, due to features including Rectified linear units and Maxout, the adversarial subspace often takes a large portion of the total input space. Once an adversarial input lies in the adversarial subspace, nearby inputs also tend to lie in it. Interestingly, this corroborates recent findings by Ilyas {\em et al.} \citep{ilyas2019adversarial} that imperceptible adversarial noise can not only be used for adversarial attacks on an already-trained network, but also as features during training. For instance, after training a network on dog images perturbed with adversarial perturbation calculated from cat images, the network can achieve a good classification accuracy on the test set of cat images. This result shows that those features (adversarial perturbations) calculated from the cat training sets, contain sufficient information for a machine learning system to make correct classification on the test set of cat images. Here, Beneficial perturbations would operate in analogous manner. We calculate those features, and store them into the beneficial perturbation bias. In this case, although the inputs data have been modified (distribution shifts of input data - information are corrupted by adversarial perturbations), the stored beneficial features have sufficient information to neutralize the effects of adversarial examples and enable the network to make correct predictions.

\subsection{Beneficial perturbations: the opposite "twins" of adversarial perturbations} Beneficial perturbations can be viewed as the opposite "twins" of  adversarial perturbations. Much research is underway on how to generate more and more advanced adversarial perturbations \citep{madry2017towards, kurakin2016adversarial,szegedy2013intriguing,narodytska2016simple,goodfellow2014explaining,kurakin2016adversarial} to fool the more and more sophisticated machine learning systems. However, there is a little research \citep{9356334} on how to generate beneficial perturbations and  possible applications of  beneficial perturbations. For example,  Wen {\em et al.} \citep{9356334} have demonstrated that beneficial perturbations can largely eliminate catastrophic forgetting (training the same neural network on a new task would destroy the knowledge learned from the old tasks) on subsequent tasks.

{\small
\bibliographystyle{ieee_fullname}
\bibliography{egbib}
}

\section{Supplementary}

\subsection{Influence of adversarial perturbation budget}
\begin{figure}[htb]
	\begin{center}
		\includegraphics[width=0.7\linewidth]{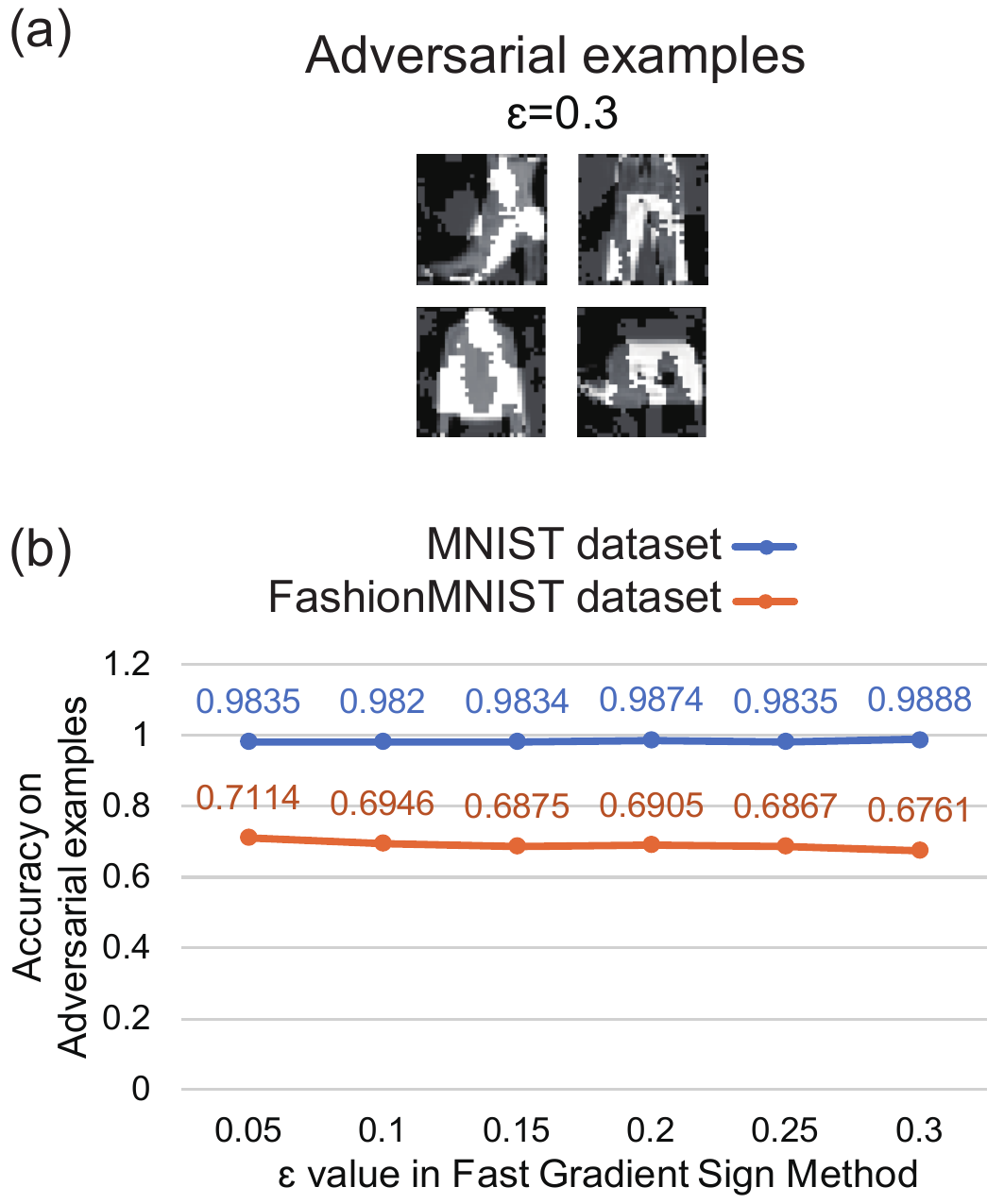}
		\caption{(a) Adversarial example with high adversarial perturbation budget ($\epsilon$ = 0.3). (b) Test accuracy on adversarial examples after training BPN only on clean examples from MNIST (Blue) or FashionMNIST (Orange) datasets. }
		\label{fig:adversarial_bugets}
	\end{center}
\end{figure}

 The higher the adversarial perturbation budget, the higher the chance it can successfully attack a neural network. However, attacks with higher adversarial perturbation budgets are easier to detect by a program or by humans. For example, $\epsilon =0.3$ (Fig.~\ref{fig:adversarial_bugets}a) represents very high noise, which makes FashionMNIST images difficult to classify, even by humans. But the distribution differences between the adversarial examples and clean examples are so large that they can be easily captured by defense programs. Thus, $\epsilon \leq 0.15$  is a good attack since the differences caused by adversarial perturbations are too small to be detected by most defense programs. For small adversarial perturbations (Fig.~\ref{fig:adversarial_bugets}b $\epsilon \leq 0.15$), by just training on clean images, BPN achieves moderate robustness against adversarial examples with negligible costs. Thus, it is really beneficial to adapt our method for companies with modest computation power, who still want to achieve moderate robustness against adversarial examples.

\subsection{Future Beneficial Perturbations research} In this research, we use one of the classical method (FGSM) available in the adversarial perturbations world to generate beneficial perturbations. We demonstrated that using beneficial perturbations can effectively defend adversarial examples by neutralizing the effects of adversarial perturbations of data samples. More research could be done to improve BPN - {{\bf(1)} Updating rules of Beneficial perturbations:} other than FGSM implemented in this paper, one could use other methods (e.g., PGD) to update the beneficial perturbation bias. As a consequence, BPN might be more robust to various kinds of adversarial examples. {{{\bf (2)} Network structure for storing and generating beneficial perturbations:}} We used beneficial perturbation biases to store and generate the beneficial perturbations. The structure of the beneficial perturbation biases is the same as normal bias. One might further optimize the storage of beneficial perturbations beyond fully connected layer biases.

\end{document}